\documentclass{article}
\usepackage{stywhispers,amsmath,epsfig}

\usepackage{microtype}
\usepackage{graphicx}
\usepackage{subfigure}
\usepackage{booktabs} 

\usepackage{hyperref}

\usepackage{amsmath}
\usepackage{amssymb}
\usepackage{mathtools}
\usepackage{amsthm}
\usepackage{qtree}

\usepackage[capitalize,noabbrev]{cleveref}

\theoremstyle{plain}

\theoremstyle{definition}

\theoremstyle{remark}

\title{Neural Network Learning of Chemical Bond Representations in Spectral Indices and Features}

\name{Bill Basener}
\address{University of Virginia, School of Data Science \\
Charlottesville, VA \\
wb8by@virginia.edu}

\begin{document}
%
\maketitle
\begin{abstract}
In this paper we investigate neural networks for classification in hyperspectral imaging with a focus on connecting the architecture of the network with the physics of the sensing and materials present.  Spectroscopy is the process of measuring light reflected or emitted by a material as a function wavelength.  Molecular bonds present in the material have vibrational frequencies which affect the amount of light measured at each wavelength.  Thus the measured spectrum contains information about the particular chemical constituents and types of bonds.  For example, chlorophyll reflects more light in the near-IR rage (800-900nm) than in the  red (625-675nm) range, and this difference can be measured using a normalized vegetation difference index (NDVI), which is commonly used to detect vegetation presence, health, and type in imagery collected at these wavelengths.  In this paper we show that the weights in a Neural Network trained on different vegetation classes learn to measure this difference in reflectance.  We then show that a Neural Network trained on a more complex set of ten different polymer materials will learn spectral 'features' evident in the weights for the network, and these features can be used to reliably distinguish between the different types of polymers.  Examination of the weights provides a human-interpretable understanding of the network.
\end{abstract}


\section{Introduction}
\label{introduction}
\begin{figure}[ht]
\vskip 0.1in
\begin{center}
\centerline{\includegraphics[width=\columnwidth]{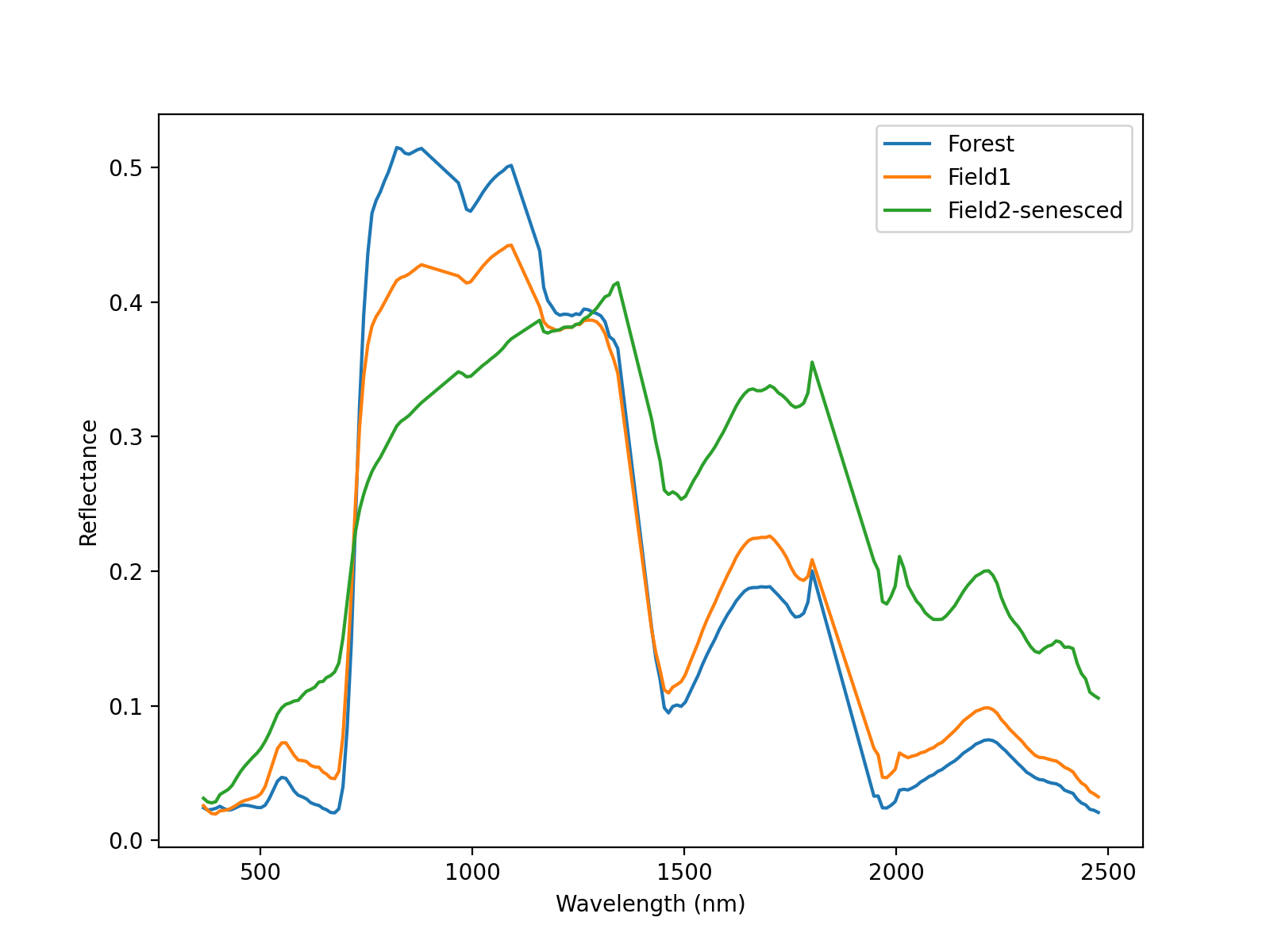}}
\caption{Spectra for three classes of vegetation.  Note the steep rise around 750nm, which is the result of chlorophyll in the vegetation, and that the scenesed (green plot) vegetation has a less prominent rise, indicating lower amounts of chlorophyll.}
\label{VegClassMeans}
\end{center}
\vskip -0.2in
\end{figure}

Hyperspectral imaging is a digital imaging technology in which each pixel contains not just the usual three visual colors (red, green, blue) but many, often hundreds, of wavelengths (or bands) of light enabling spectroscopic information about the materials located in the pixel~\cite{SchottBook2007}.  A common range of wavelengths covers the VNIRSWIR (visible, near-infrared, short-wave infrared) from about 450nm to 2500nm, where the visible red, green, and blue colors correspond approximately to 650nm, 550nm, and 450nm respectively.  Each band in the image measure light in a range of about 10nm or less and the bands contiguously cover the range of wavelengths for the sensor.  In each wavelength, some percentage of light is absorbed by the material located in the pixel and some percentage is reflected back, so that the measured spectrum is a vector of percent reflectance for each band.  The absorption or reflection depends mainly on the interaction of the light at the given wavelength and molecular elements and bonds present, enabling some level spectroscopy - determining information about the materials present from the measured spectrum.  A plot of three spectra, which are the class means for imagery we use in this paper, are shown in Figure~\ref{VegClassMeans}.

Spectra can be collected from individual pixels from a hyperspectral image on a satellite~\cite{Middleton2013} or an aircraft~\cite{ForestRadiance97} in which case the material is illuminated by sunlight, or from a sensor in a laboratory setting with controlled illumination~\cite{BasenerFlynn2018}.  For outdoor settings with sunlight, atmospheric correction must be used to get at-surface reflectance from the at-sensor measured radiance~\cite{Perkins2012}.  We will be working with both overhead-collected data and laboratory-collected data, all of which is in units of percent reflectance.

There are known computations using the reflectance values in specific ranges of wavelengths that measure particular physical properties for materials on the ground are known as \textit{spectral indices}.  These are determined from variation in percent reflectance that is consistent for a particular material or varies in a consistent way with variation of a physical property.  These consistent variation in reflectance at specific wavelengths are called \textit{spectral features}.

A standard spectral index is the normalized difference vegetation index (NDVI), which measures a known feature called the 'red edge' for vegetation identification and health/abundance assessment.  This red edge is observable as the steep rise around 700nm in Figure~\ref{VegClassMeans}.  The NDVI is defined as:
\[
\textrm{NDVI}=\frac{NIR-R}{NIR+R}
\]
where $R$ (for red)is the percent reflectance across bands around 650nm and NIR (for near-infrared) is percent reflectance for bands around 750-900nm.

This is a standard measurement for vegetation, particularly common with multispectral imagery which has few bands than hyperspectral and with each band spanning a broader range of wavelengths.  For example, The NASA Landsat multispectral imagery satellite program~\cite{Wulder2019} offers an NDVI image (which has the NDVI value for each pixel) as a standard product.  For Landsat 8 and 9 NDVI is computed using 850-880nm as the range for the NIR band and 640-670nm as the Red band, while Landsat 4-7 use approximately 760-900nm for the NIR band and 630-690nm for the Red band.  For a hyperspectral image, the Red and NIR reflectance can be measued using reflectance for a single band in one of these ranges or the average reflectance over some set of bands for each range.

In this paper we demonstrate that a neural network can effectively learn the NDVI.  We show that the feature is evident in the network's weights, and we also show that the network separates the data into groups similar to separating by NDVI values.  We then show a neural network learning numerous indices around features for a complex set of ten different polymer materials.  For each dataset, the data is split 50/50 with stratification into training and testing sets.  The network trained on the polymer training data has 99.98\% accuracy on the test set, and the network trained on the vegetation training data has 99.92\% accuracy on its test set.

\section{Data}
\begin{figure}[ht]
\begin{center}
\centerline{\includegraphics[width=\columnwidth]{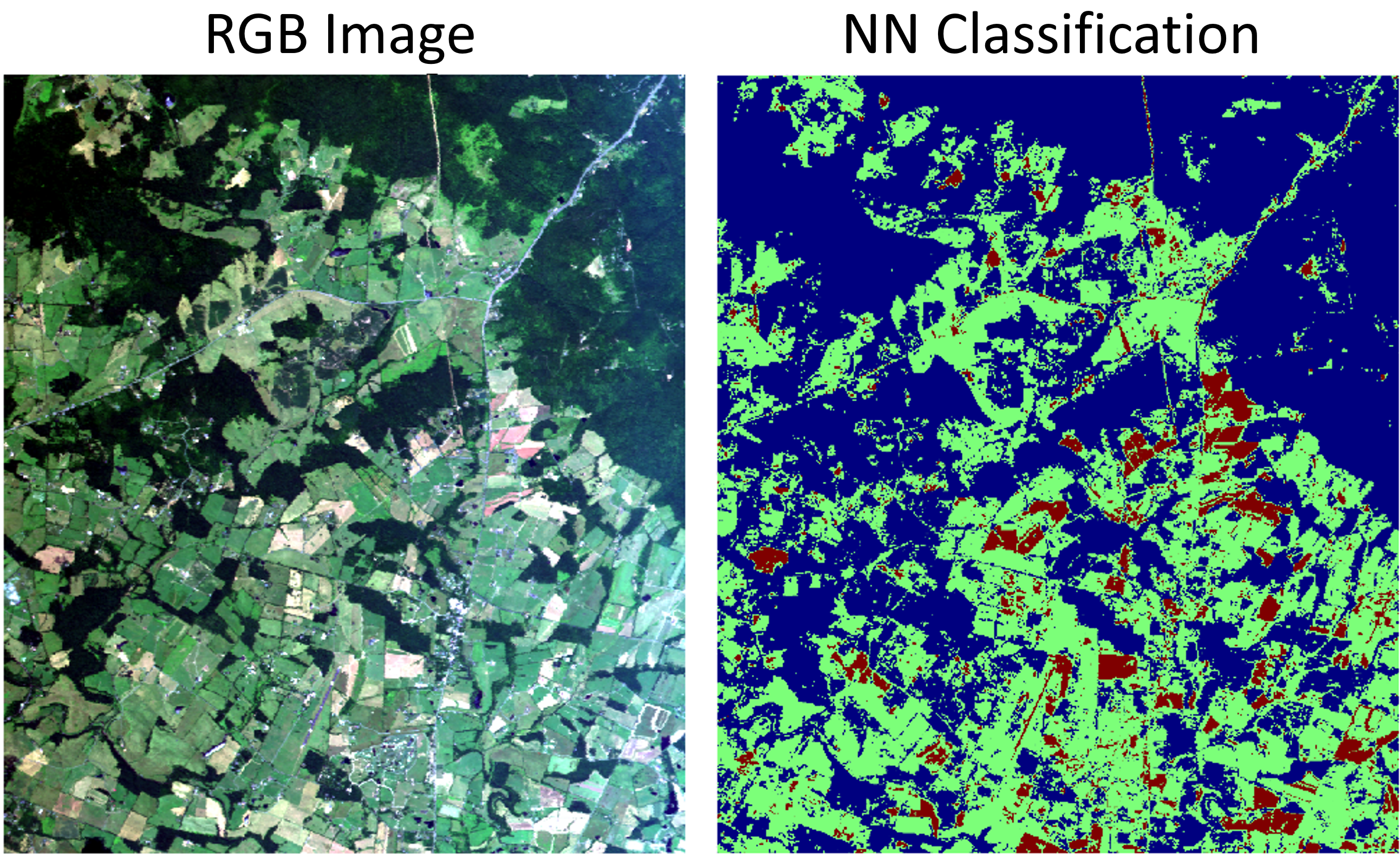}}
\caption{(Left) An RGB color image from the AVIRIS hyperspectral image.  (Right) Classification results on the image using the NN.  The classes are dark green trees, light green field, and a tan scenesed field colored blue, green, and red respectively in the classification images.}
\label{AVRISrgb}
\end{center}
\vskip -0.2in
\end{figure}

The hyperspectral image we use to collect classes of vegetation data is shown in Figure~\ref{AVRISrgb}.  We selected three rectangular regions from this image, one on dark green vegetation (trees), one on light green vegetation (a field), and one on a senesced (dry tan) field, and showed the result of classification using the neural network.
The image has 181 bands a total of 560,000 pixels.  There are 2,580 pixels labeled for the forest class has, 168 the field1 class, and 104 pixels for the field2\_senesced class.  Senescence is a process that occurs in vegetation during which chlorophyll content decreases and the vegetation looses its visual green color.  This may be a seasonal variation or an indication that the vegetation is stressed for example from a lack of water.

\begin{figure}[ht]
\begin{center}
\centerline{\includegraphics[width=0.9\columnwidth]{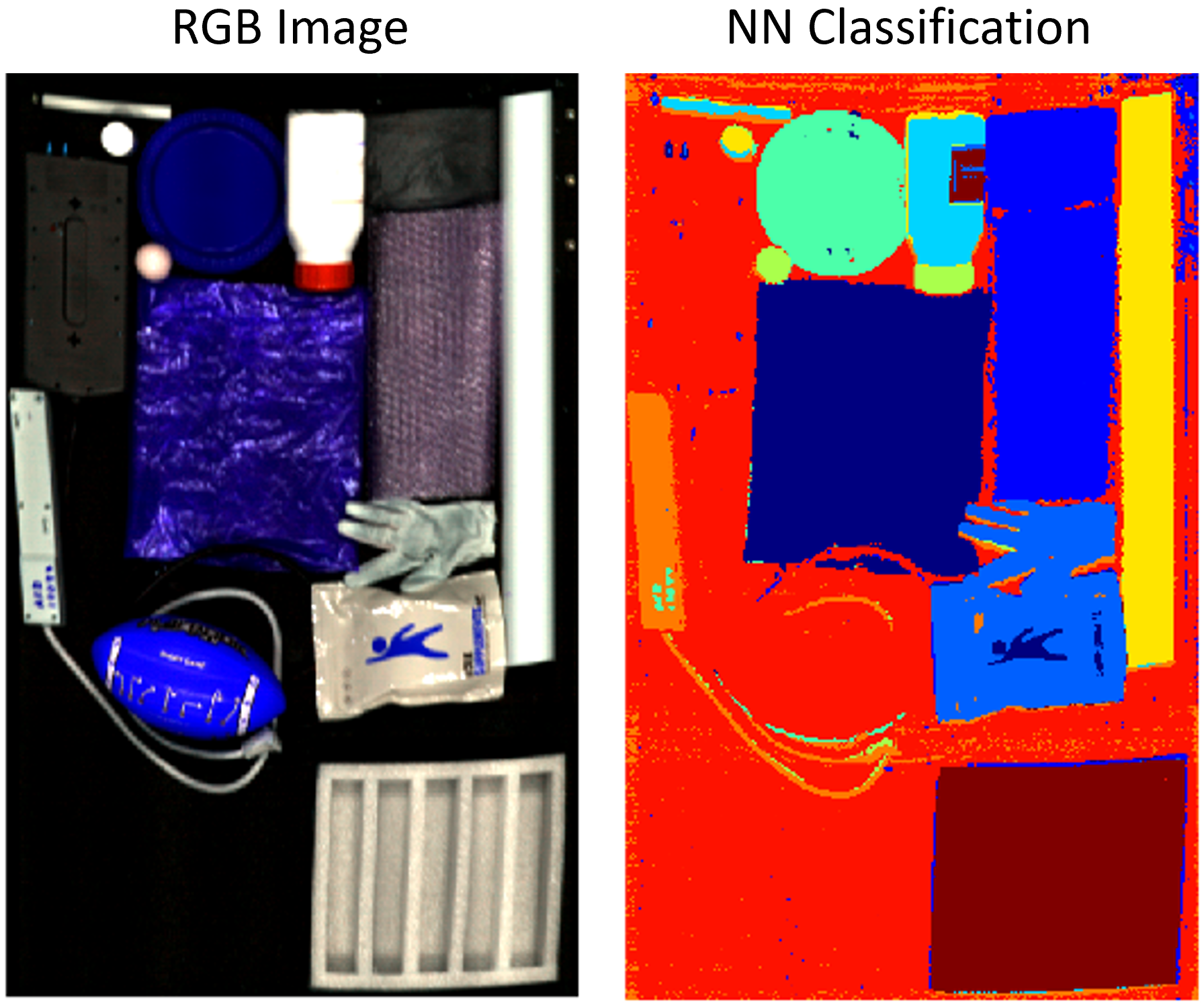}}
\caption{(Left) An RGB color image from the SPECIM hyperspectral polymer image.  (Right) Classification results on the image using the NN.}
\label{POLYrgb}
\end{center}
\vskip -0.2in
\end{figure}

The hyperspectral image we use to collect polymer spectra is shown in Figure~\ref{POLYrgb}, with an RGB representation on the left and the NN classification on the right.  This image was collected using a SPECIM spectrometer in a lab, with controlled illumination and materials on a moving table similar to~\cite{BasenerFlynn2018}.  The image has 452 bands and a total around 320,000 pixels.  The number of pixels that were selected for each class are shown in Table~\ref{PolyPixelCounts}.
\begin{table}[t]
\caption{The number of pixels per class.}
\label{PolyPixelCounts}
\begin{center}
\begin{small}
\begin{sc}
\begin{tabular}{lcccr}
\toprule
Class Name & Number of Pixels \\
\midrule
red\_bubble\_wrap & 3672     \\
clear\_bubble\_wrap & 3003     \\
glove\_loc & 1200     \\
medicine\_bottle & 1360     \\
red\_lid & 2958     \\
ping\_pong\_ball & 225     \\
pvc\_pipe & 5415     \\
pvc\_extension\_plug & 1818     \\
inflatable\_football & 3196     \\
foam\_packaging & 8900     \\
\end{tabular}
\end{sc}
\end{small}
\end{center}
\vskip -0.1in
\end{table}

\section{Methods}
For each dataset, the data is split 50/50 with stratification into training and testing sets. For each training set, we constructed a neural network with a single hidden layer of 128 ReLu neurons, followed by a 20\% dropout layer, and the a softmax classification layer with the number of neurons equal to the number of classes.  We trained the network for 50 epochs using Keras for Tensorflow.  Each network was applied to the test data to obtain test accuracy and also to the full image to investigate how the network acts more generally.

This is a relatively simpler neural network, and we used a relatively large dropout rate which corresponds to Bayesian model averaging on the neurons.  This simple network architecture makes the model interpretable so that we can observe the learning of spectra features in the weights.  The high drouput rate provides sufficient variation to get a robust measure of Bayesian uncertainty about the weights for each class.

\section{Results}
\begin{figure}[ht]
\begin{center}
\centerline{\includegraphics[width=\columnwidth]{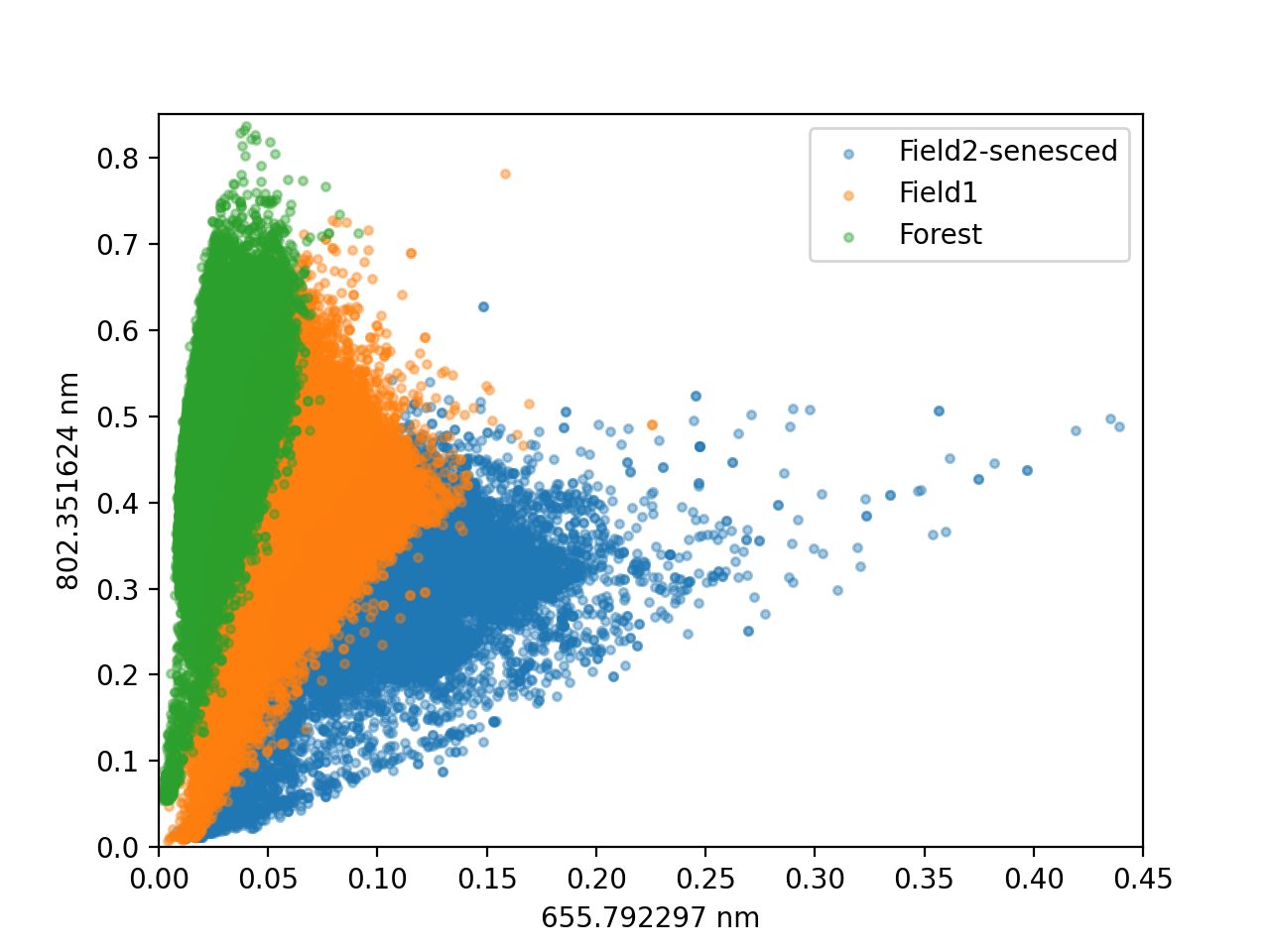}}
\caption{A scatterplot of the spectra from the AVIRIS image with colors showing the labels from the neural network.}
\label{Avris_NN_Classes_Scatterplot}
\end{center}
\vskip -0.2in
\end{figure}

\begin{figure}[ht]
\begin{center}
\centerline{\includegraphics[width=\columnwidth]{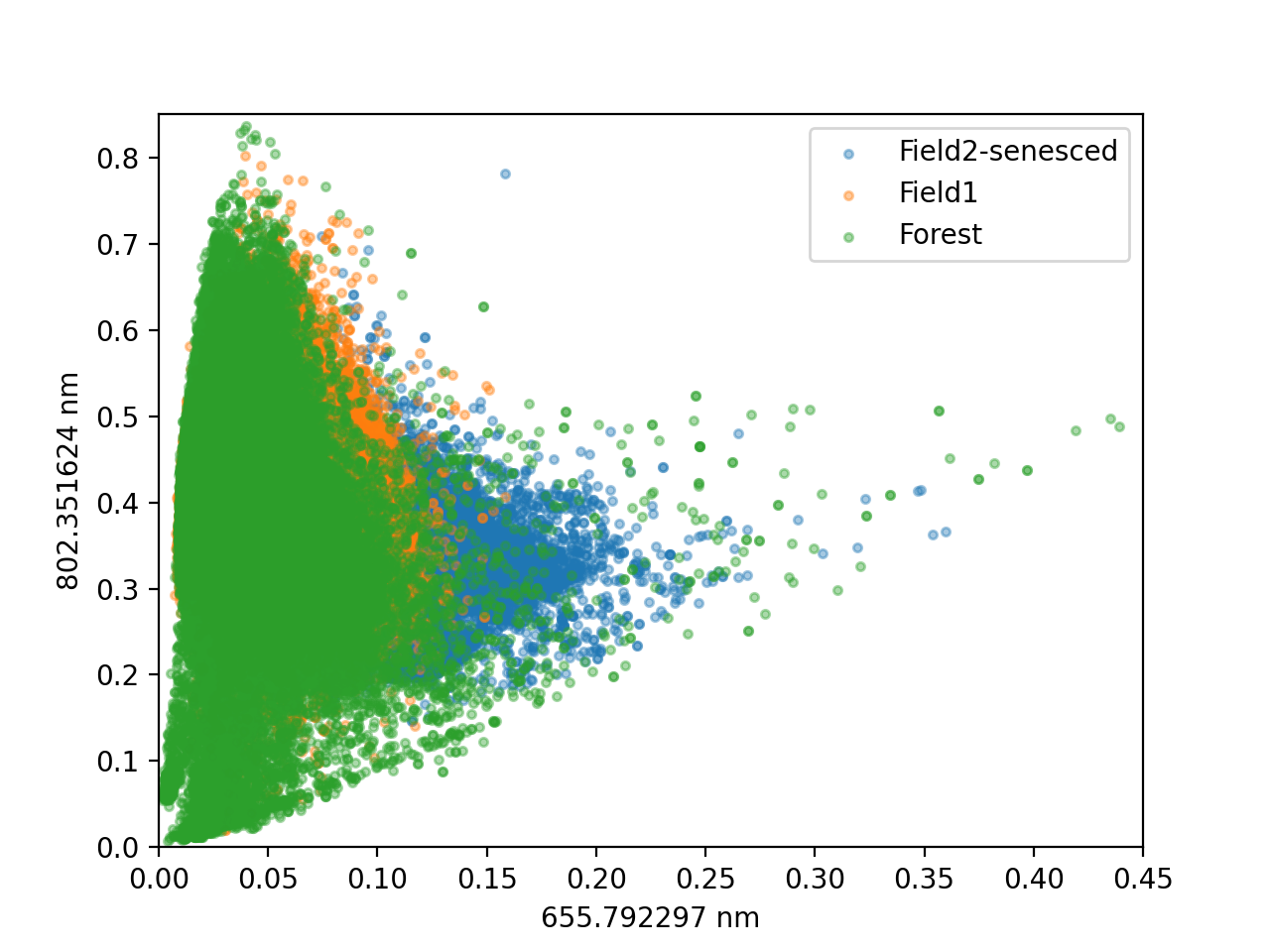}}
\caption{A scatterplot of the spectra from the AVIRIS image with colors showing the labels from Linear Discriminant Analysis.}
\label{Avris_LDA_Classes_Scatterplot}
\end{center}
\vskip -0.2in
\end{figure}

The network trained on the polymer training data has 99.98\% accuracy on the test set, and the network trained on the vegetation training data has 99.92\% accuracy on its test set.  This is of course excellent accuracy, and the rest of this section is devoted to interpreting the network as a means of leaning spectral indices and features; first leanding NDVI for vegetation and then applications for the polymers.  Figure~\ref{Avris_NN_Classes_Scatterplot} shows a scatterplot of the spectra from the AVIRIS image with reflectance in a red band (656nm) on the $x$-axis and reflectance in an infrared (802nm) band as the $y$-axis, with colors showing the labels from the neural network.  For comparison, Figure~\ref{Avris_LDA_Classes_Scatterplot} shows a scatterplot of this data on the same bands, but with colors showing labels from an LDA classifier that was trained on the same data as the neural network.

To see why the scatterplot in Figure~\ref{Avris_NN_Classes_Scatterplot} represents separation by NDVI, observe from the formula for NDVI that NDIV is a number in $[-1,1]$, and that for a fixed value of NDVI the relationship bewteen the NIR and R values is
\[
NIR=R\frac{\textrm{NDVI}+1}{1-\textrm{NDVI}}.
\]
This is a linear function, and the slope $\frac{\textrm{NDVI}+1}{1-\textrm{NDVI}}$ is zero for $\textrm{NDVI}=-1$ and monotonically increasing as NDVI increases.  Thus, a given NDVI value corresponds to a line through the origin in the plane whose $x$-axis is reflectance in a red band and whose $y$-axis is reflectance in infrared, and spectra with larger NDVI values will lie above this line.   The points are colored according to the classes from the network, and observe the linear separation in Figure~\ref{Avris_NN_Classes_Scatterplot} shows that the network is separating pixels with the same geometry as NDVI when viewed in these bands.  For comparison, the classification shown by LDA in Figure~\ref{Avris_LDA_Classes_Scatterplot} does not follow this geometry.

\begin{figure}[ht]
\begin{center}
\centerline{\includegraphics[width=\columnwidth]{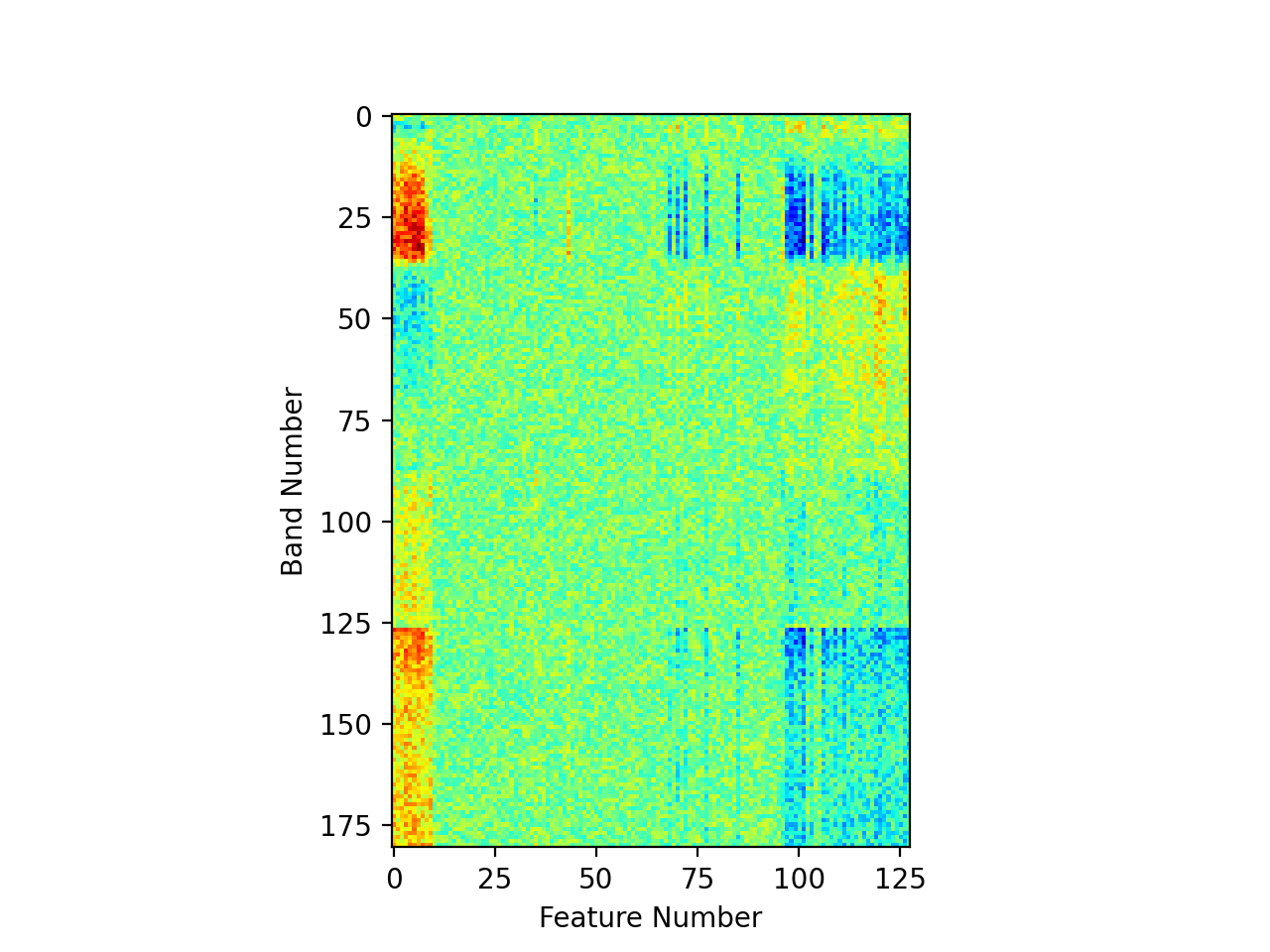}}
\caption{The weights in the hidden layer of the neural network trained on the vegetation data.  The band input into each neuron is on the y-axis and the neurons are along the x-axis sorted by the weights assigned to each neuron by the final softmax classification layer.}
\label{vegNNweights}
\end{center}
\vskip -0.2in
\end{figure}

\begin{figure}[ht]
\begin{center}
\centerline{\includegraphics[width=\columnwidth]{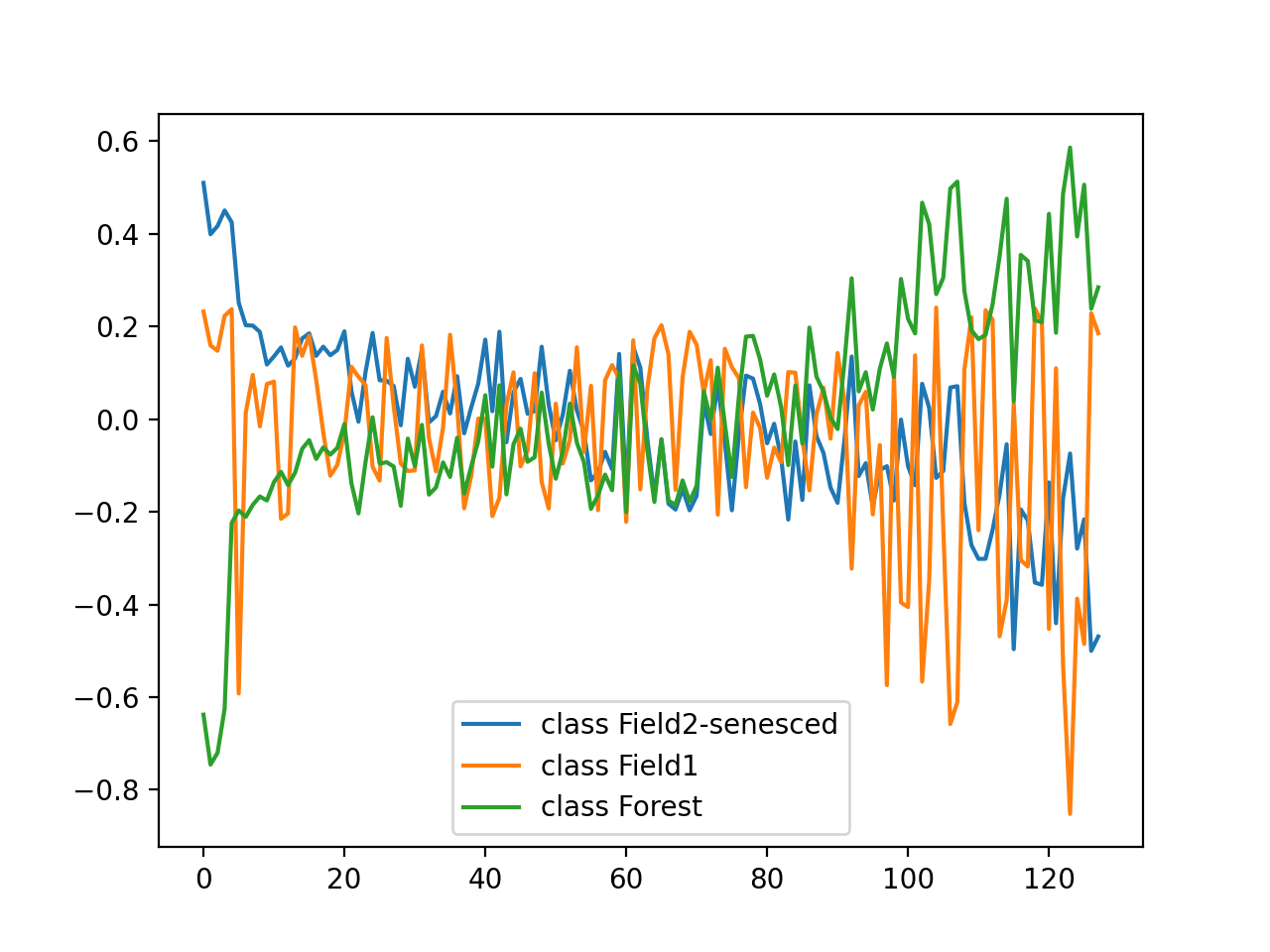}}
\caption{The weights from the hidden layer, colored by the class assignment by the neuron in the final layer.}
\label{vegNNweightsFinal}
\end{center}
\vskip -0.2in
\end{figure}

The matrix of weights for the neural network trained on the vegetation data is shown in Figure~\ref{vegNNweights}.  The band input into each neuron is on the vertical axis and the neurons are along the horizontal axis sorted by the weights assigned to each neuron by the final softmax classification layer.  These are sorted approximately based on how they are used for classification in the final layer, with the weights from the hidden layer to final classification layer shown in Figure~\ref{vegNNweightsFinal}.  The horizontal axis these both of these figures corresponds to neurons in the hidden layer, and they are provided in the same order.  The first approximately 10 neurons are weighed highly for the softmax classification neuron for the Forest class, the middle group had higher weights for the Field1 class, and the last approximately 25 had the highest weights as inputs into the softmax classification neuron for the Field2-senesced class.

\begin{figure}[ht]
\begin{center}
\centerline{\includegraphics[width=\columnwidth]{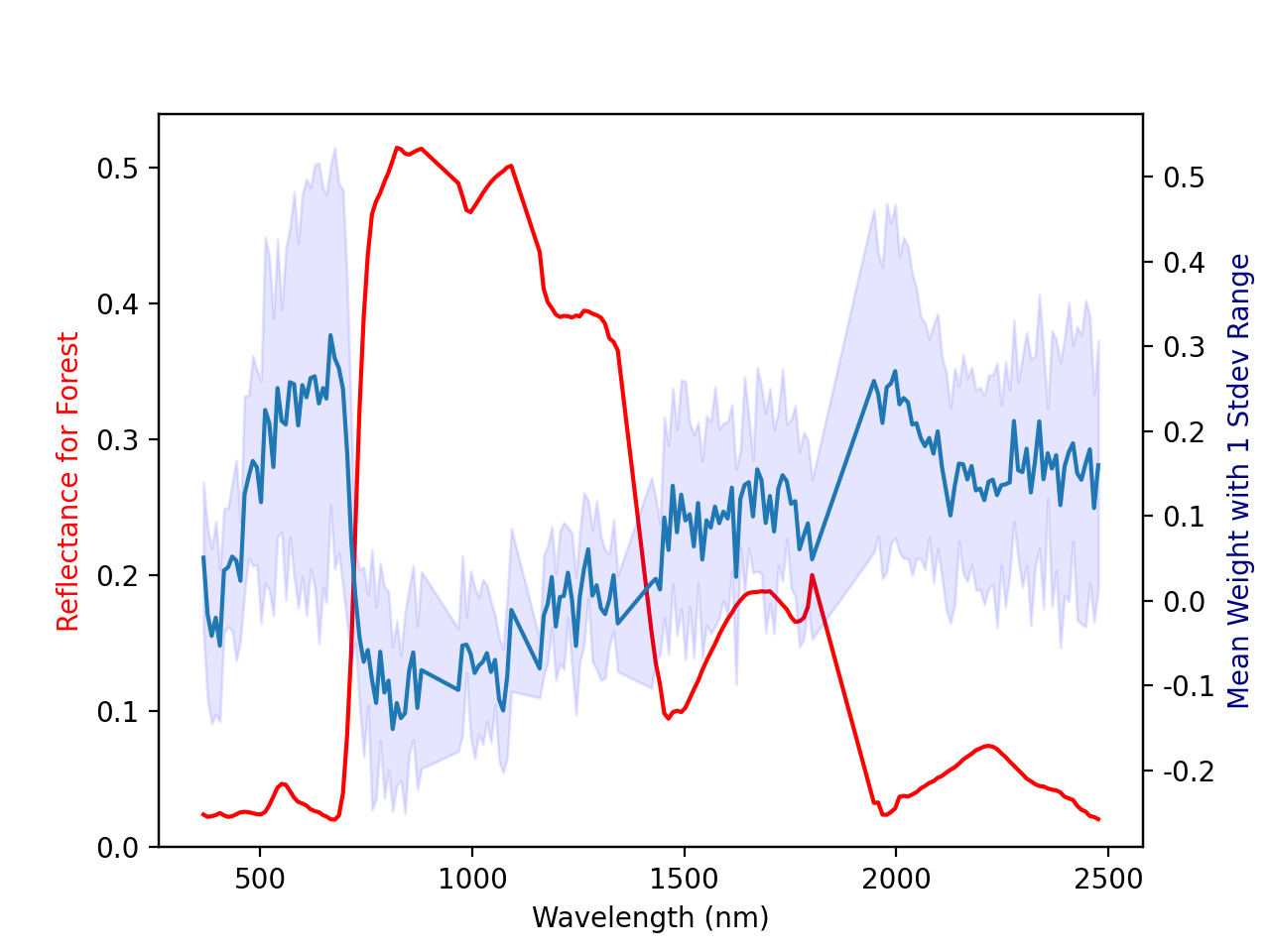}}
\caption{The mean reflectance per wavelength for the Forest class (red) along with the mean and standard deviation for weights from each wavelength going to hidden neurons associated with detecting this class.}
\label{AVIRISspectralPlot0}
\end{center}
\vskip -0.2in
\end{figure}

\begin{figure}[ht]
\begin{center}
\centerline{\includegraphics[width=\columnwidth]{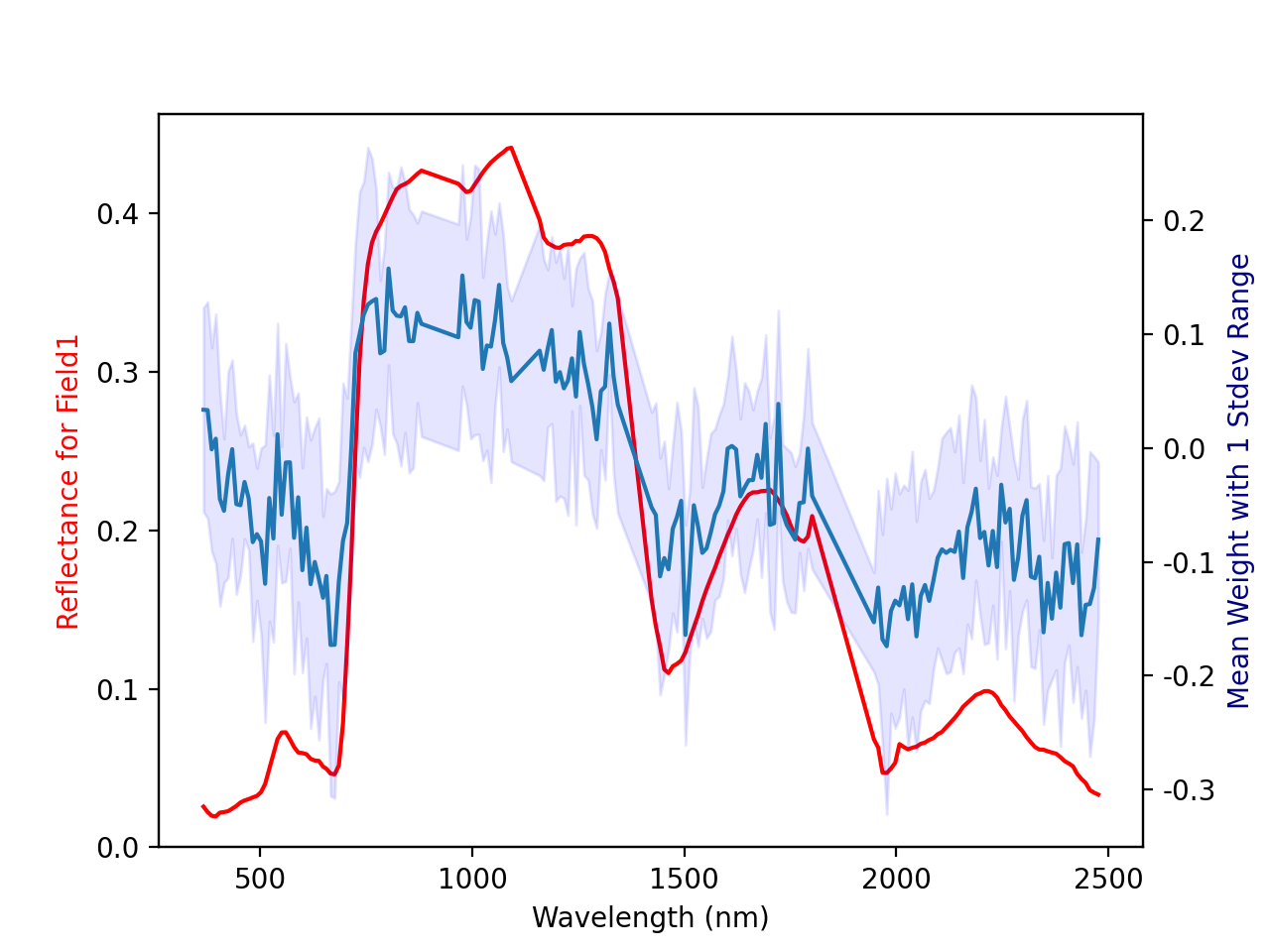}}
\caption{The mean reflectance per wavelength for the Field1 class (red) along with the mean and standard deviation for weights from each wavelength going to hidden neurons associated with detecting this class.}
\label{AVIRISspectralPlot1}
\end{center}
\vskip -0.2in
\end{figure}

\begin{figure}[ht]
\begin{center}
\centerline{\includegraphics[width=\columnwidth]{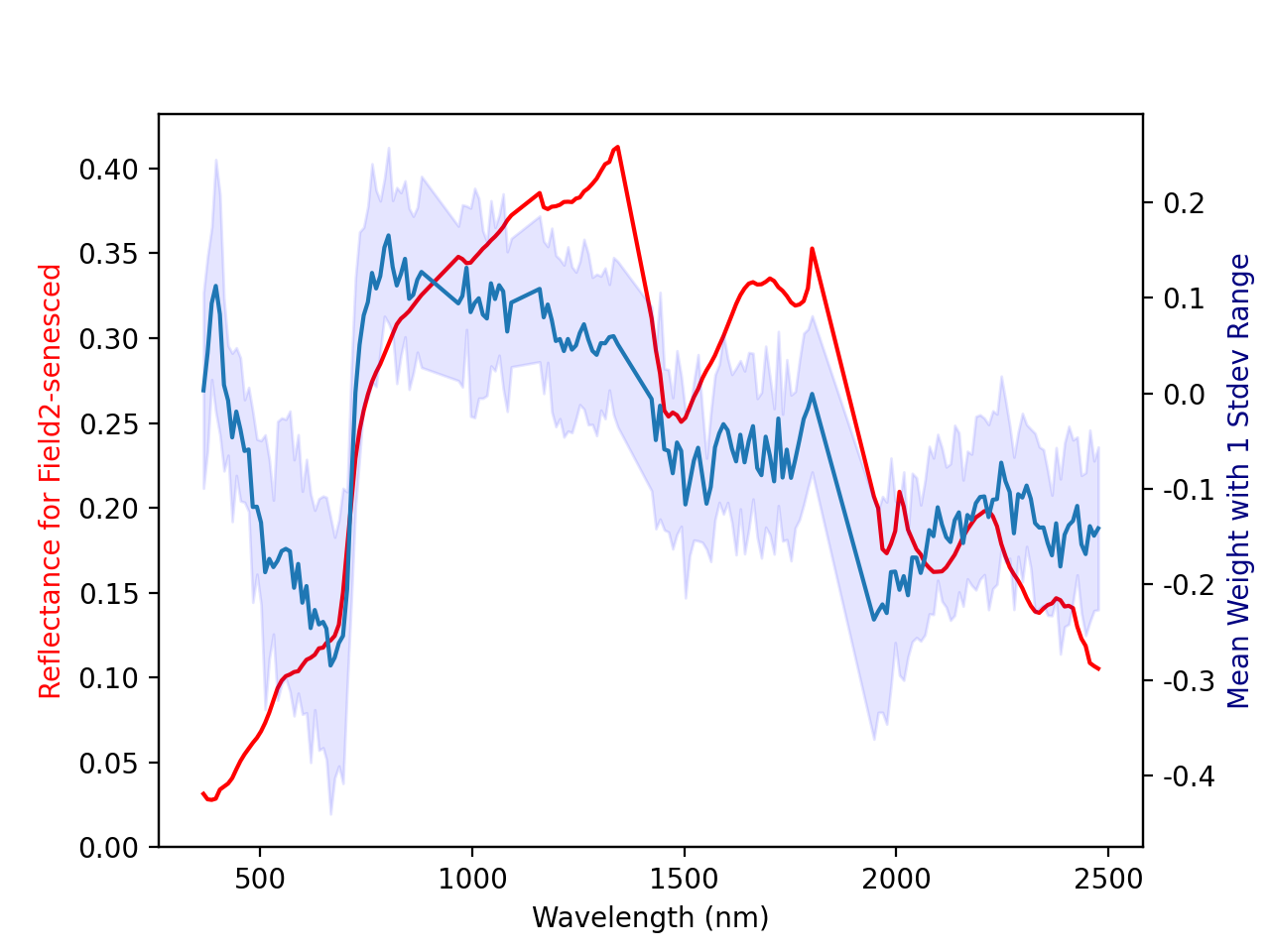}}
\caption{The mean reflectance per wavelength for the Field2-senesced class (red) along with the mean and standard deviation for weights from each wavelength going to hidden neurons associated with detecting this class.}
\label{AVIRISspectralPlot2}
\end{center}
\vskip -0.2in
\end{figure}

From the weights of the hidden layer shown in Figure~\ref{vegNNweights}, the weights on the bands for the ten neurons whose outputs have the highest weights in the softmax classification neuron for each class were selected as weights that 'learn' a representation of the respective class.  The mean and standard deviation for the inputs from each band were computet and are plotted for each group in Figures~\ref{AVIRISspectralPlot0} through~\ref{AVIRISspectralPlot2}.  In each plot, observe that the weights for the neurons have learned to measure the contrast between reflectance values on either side of the red edge as primary features for distinguishing these three classes, with increasing emphasis on bands nearest the edge.  The weights also show a learning of additional features; for example while the red edge is the strongest feature for the Forest class, the weights for the Field1 class seem to put some emphasis on visible green light (around 550nm) which is not present in the weights for the other classes.

\begin{figure}[ht]
\begin{center}
\centerline{\includegraphics[width=\columnwidth]{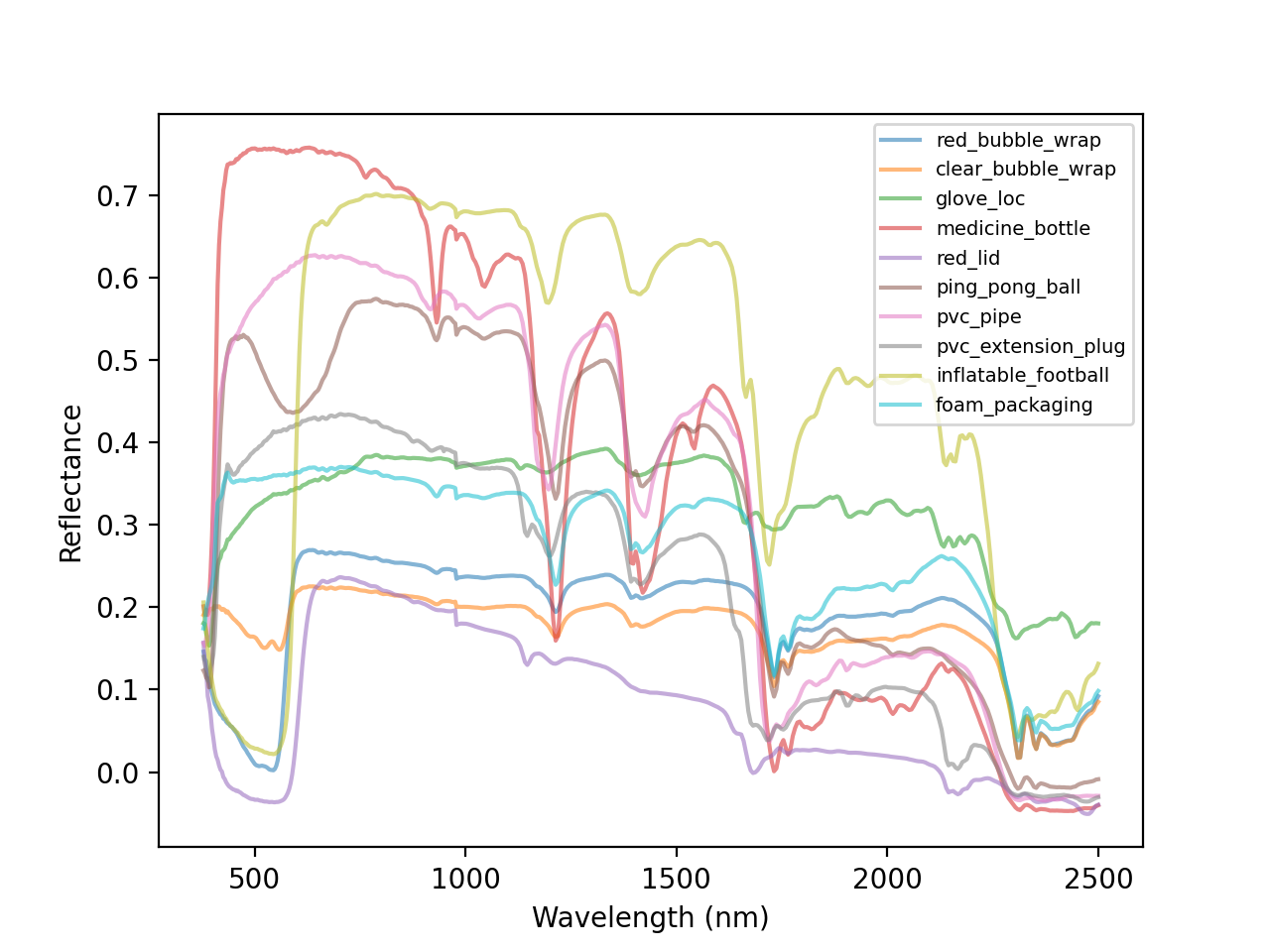}}
\caption{The mean reflectance spectrum for each polymer class.}
\label{POLYmeans}
\end{center}
\vskip -0.2in
\end{figure}

\begin{figure}[ht]
\begin{center}
\centerline{\includegraphics[width=\columnwidth]{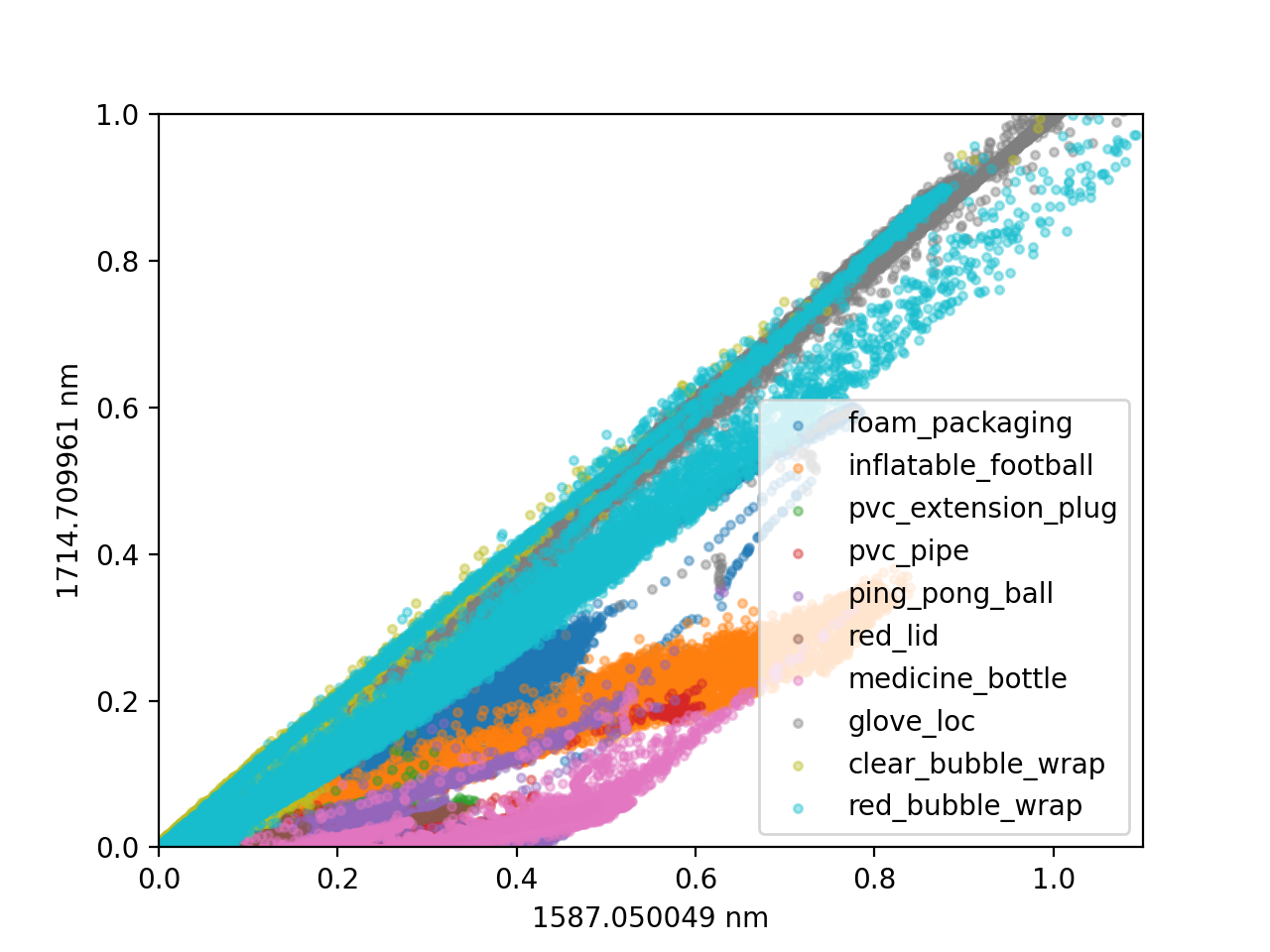}}
\caption{A scatterplot of the polymer spectra colored by labels from the neural network.}
\label{POLYscatter}
\end{center}
\vskip -0.2in
\end{figure}

A neural network, also with 128 ReLu neurons in a single hidden layer followed by a 20\% dropout layer and a final classification layer of 10 softmax classification neurons (one for each class) for the classes described in Table~\ref{PolyPixelCounts} and shown in Figure~\ref{POLYrgb}.  The mean spectrum for each class is shown in Figure~\ref{POLYmeans}, and a scatterplot of the classification results of this network are shown in Figure~\ref{POLYscatter}.  Observe that is more complexity and diversity in class means, as well as an increase in the number of classes, in comparison to the vegetation data.

\begin{figure}[ht]
\begin{center}
\centerline{\includegraphics[width=0.85\columnwidth]{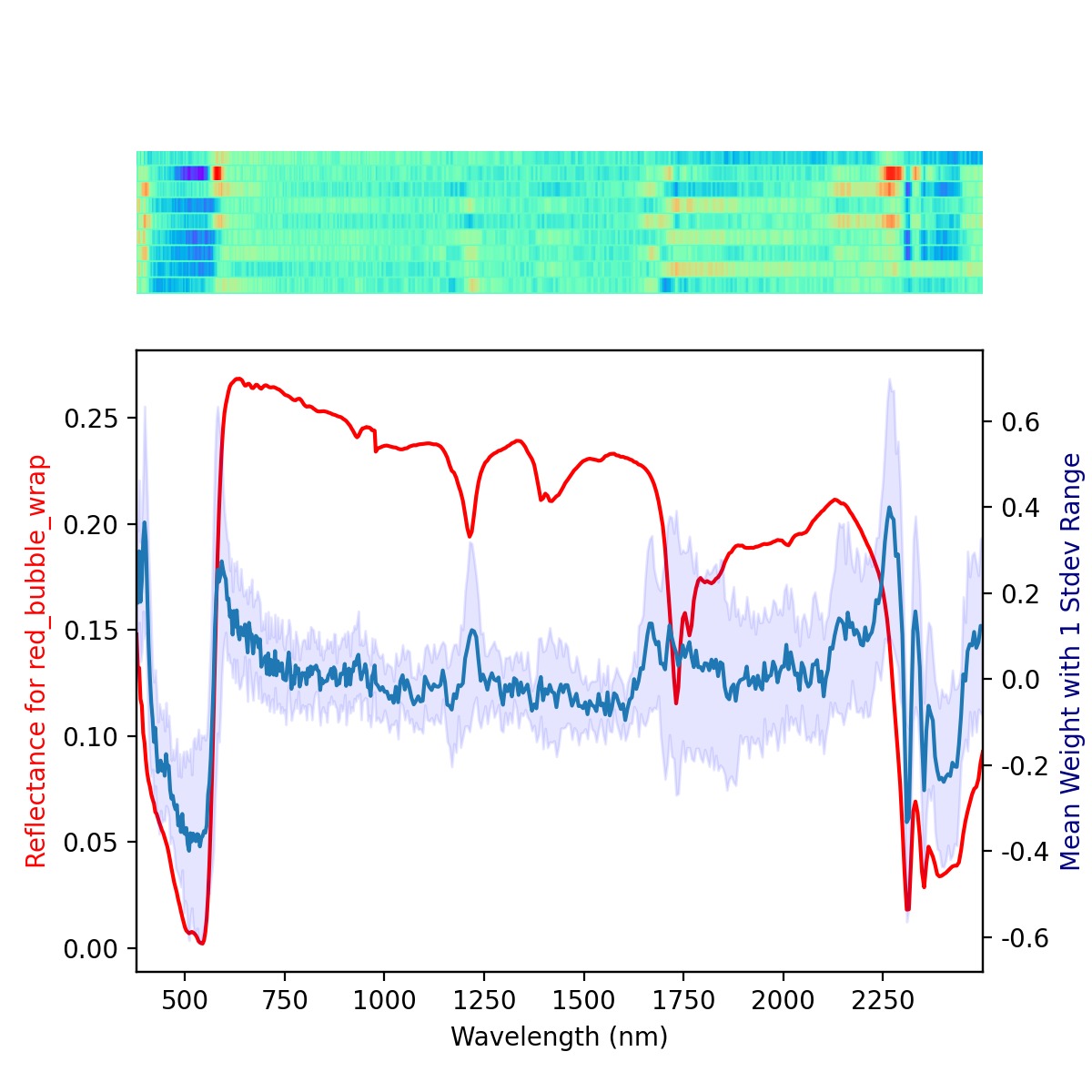}}
\vskip -0.2in
\caption{The mean reflectance for the red\_bubble\_wrap class (red) along with the associated weights.}
\label{POLYspec0}
\end{center}
\vskip -0.3in
\end{figure}

\begin{figure}[ht]
\begin{center}
\centerline{\includegraphics[width=0.85\columnwidth]{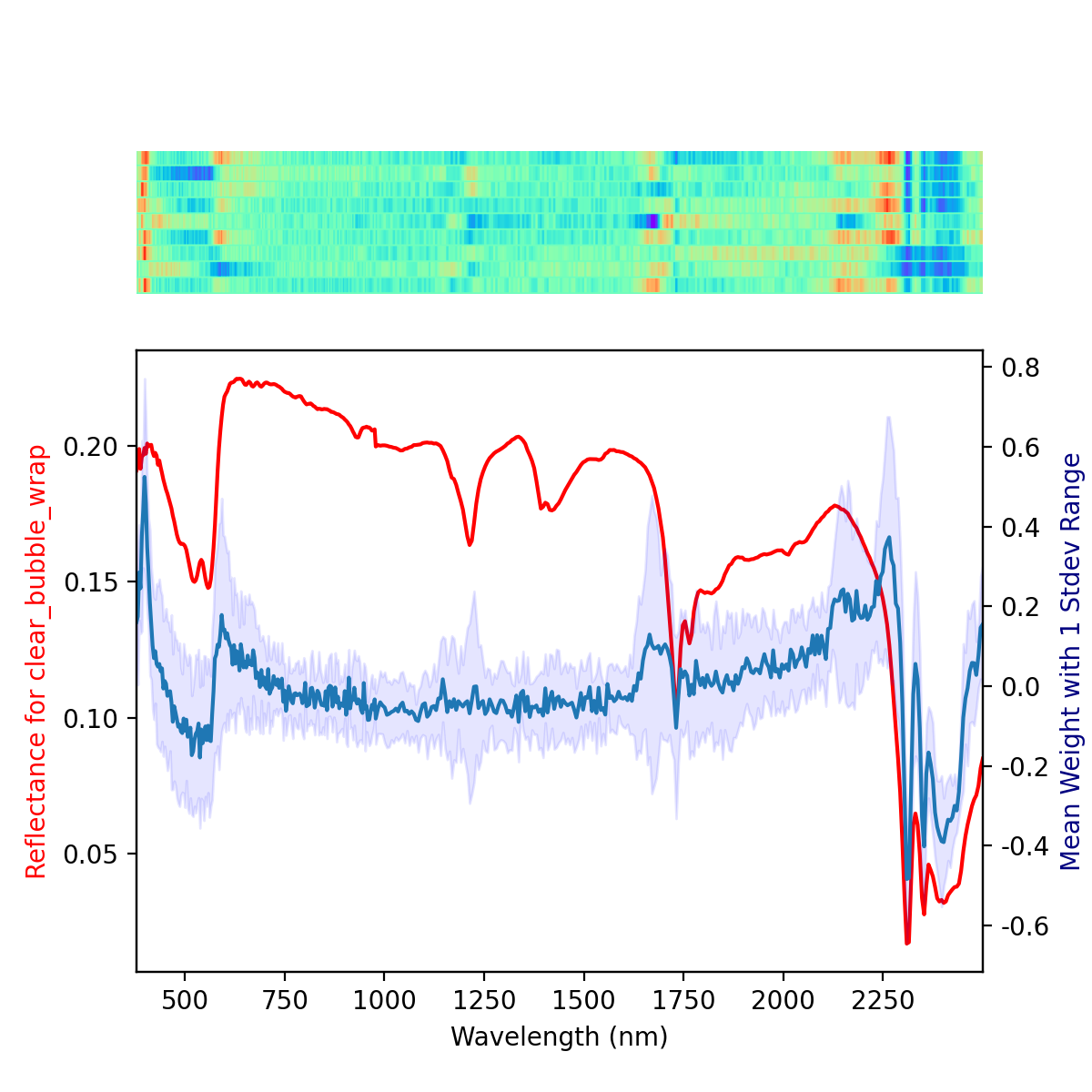}}
\vskip -0.2in
\caption{The mean reflectance for the clear\_bubble\_wrap class (red) along with the associated weights.}
\label{POLYspec1}
\end{center}
\vskip -0.3in
\end{figure}

\begin{figure}[ht]
\begin{center}
\centerline{\includegraphics[width=0.85\columnwidth]{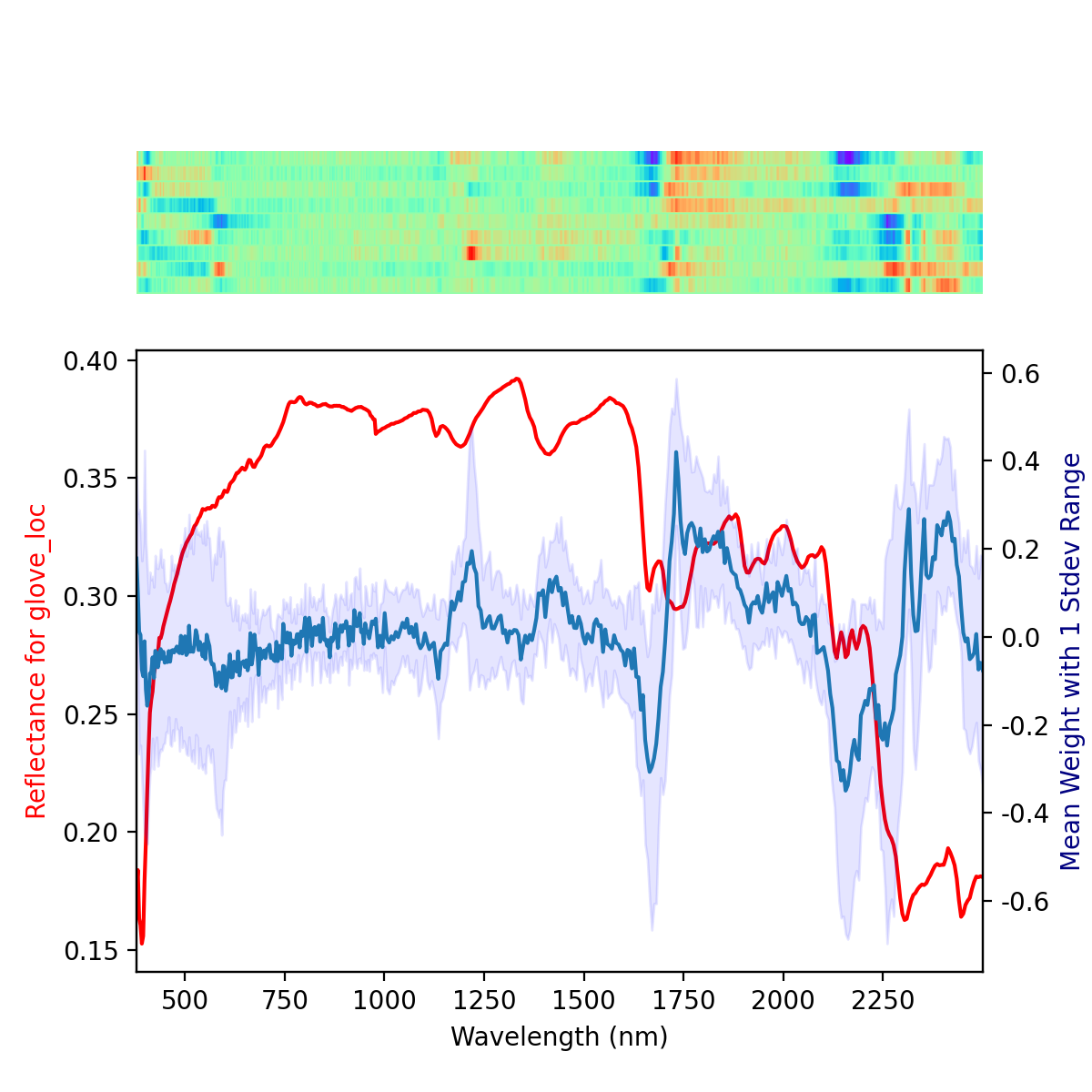}}
\vskip -0.2in
\caption{The mean reflectance for the glove\_loc class (red) along with the associated weights.}
\label{POLYspec2}
\end{center}
\vskip -0.3in
\end{figure}

\begin{figure}[ht]
\begin{center}
\centerline{\includegraphics[width=0.85\columnwidth]{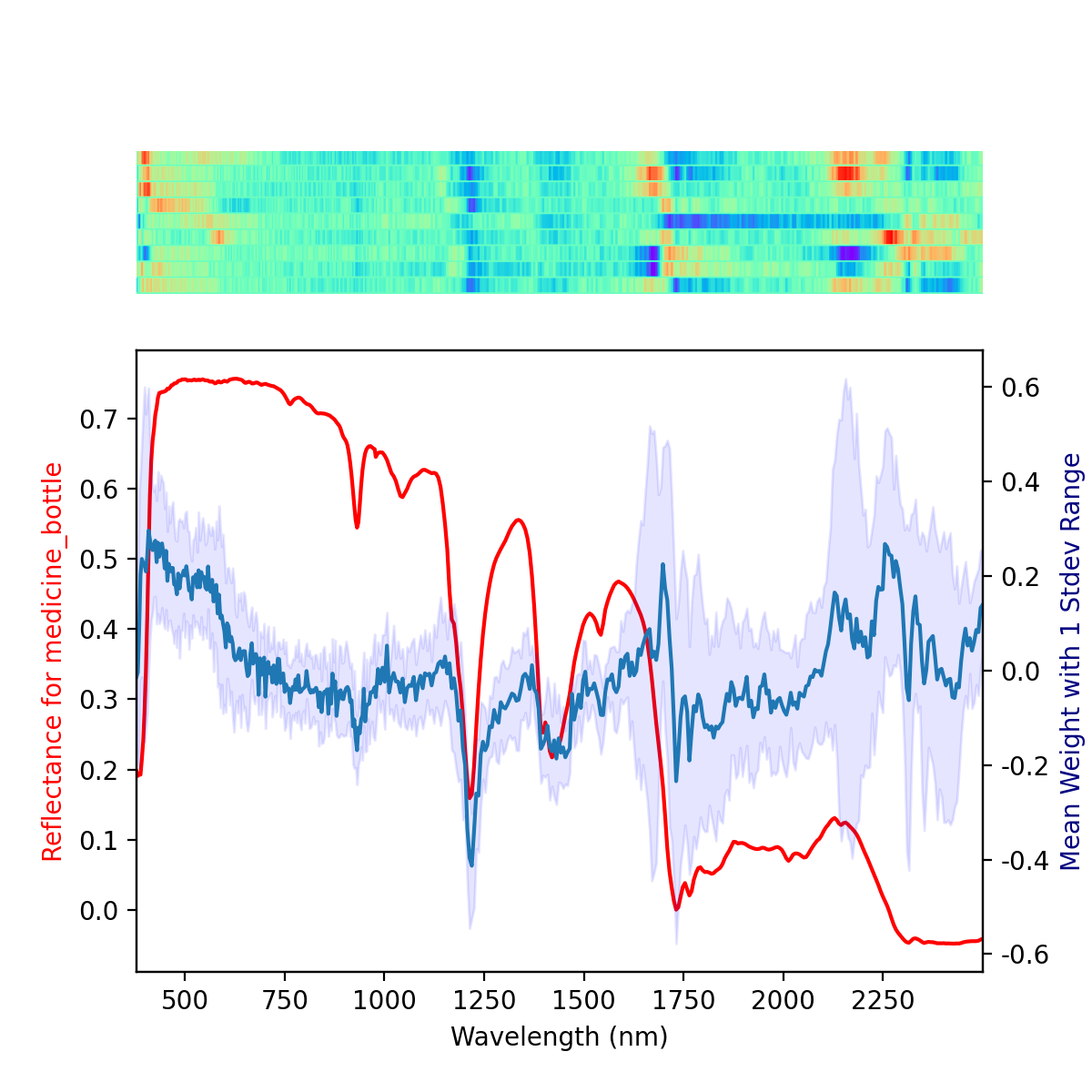}}
\vskip -0.2in
\caption{The mean reflectance for the medicine\_bottle class (red) along with the associated weights.}
\label{POLYspec3}
\end{center}
\vskip -0.3in
\end{figure}

\begin{figure}[ht]
\begin{center}
\centerline{\includegraphics[width=0.85\columnwidth]{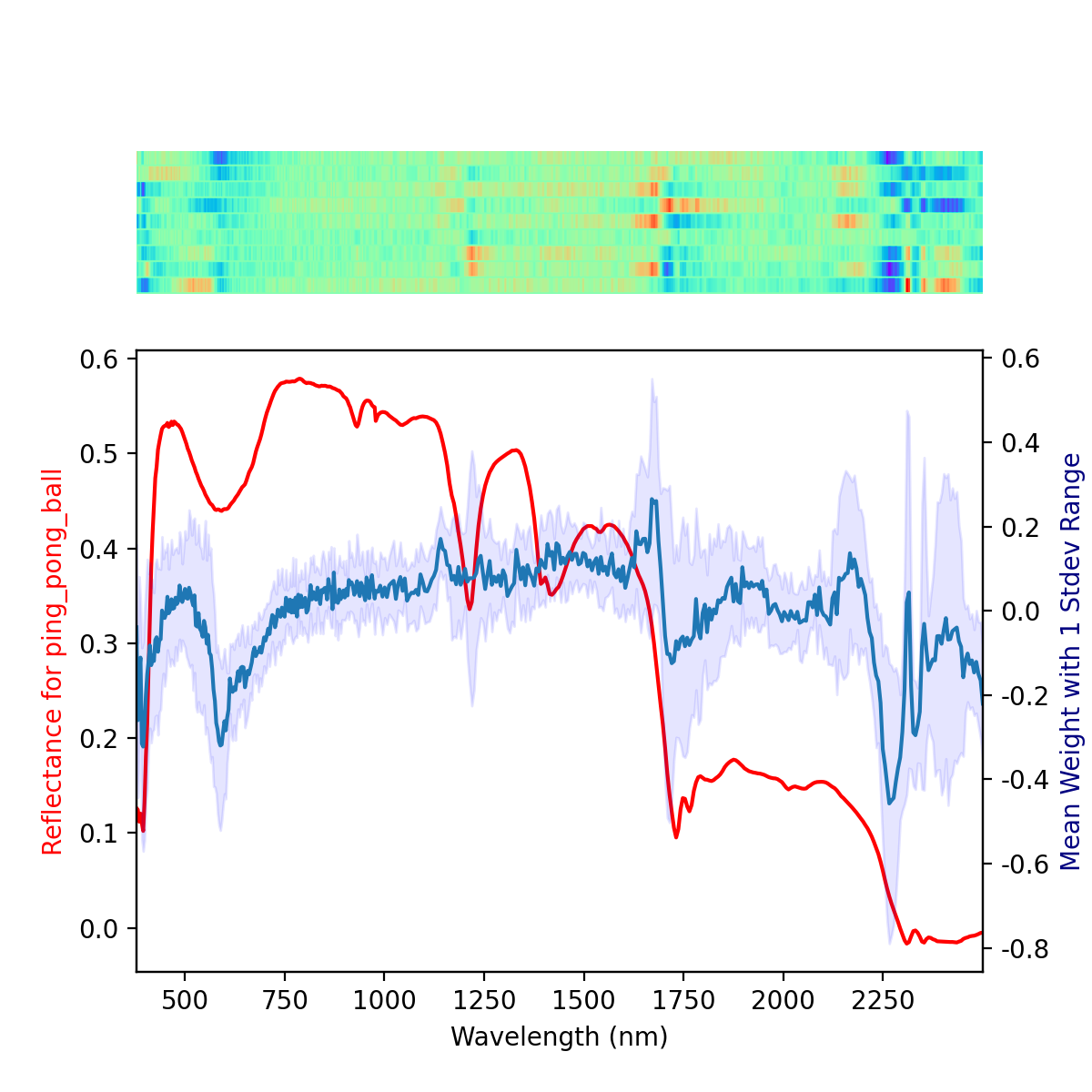}}
\vskip -0.2in
\caption{The mean reflectance for the ping\_pong\_ball class (red) along with the associated weights.}
\label{POLYspec4}
\end{center}
\vskip -0.3in
\end{figure}

\begin{figure}[ht]
\begin{center}
\centerline{\includegraphics[width=0.85\columnwidth]{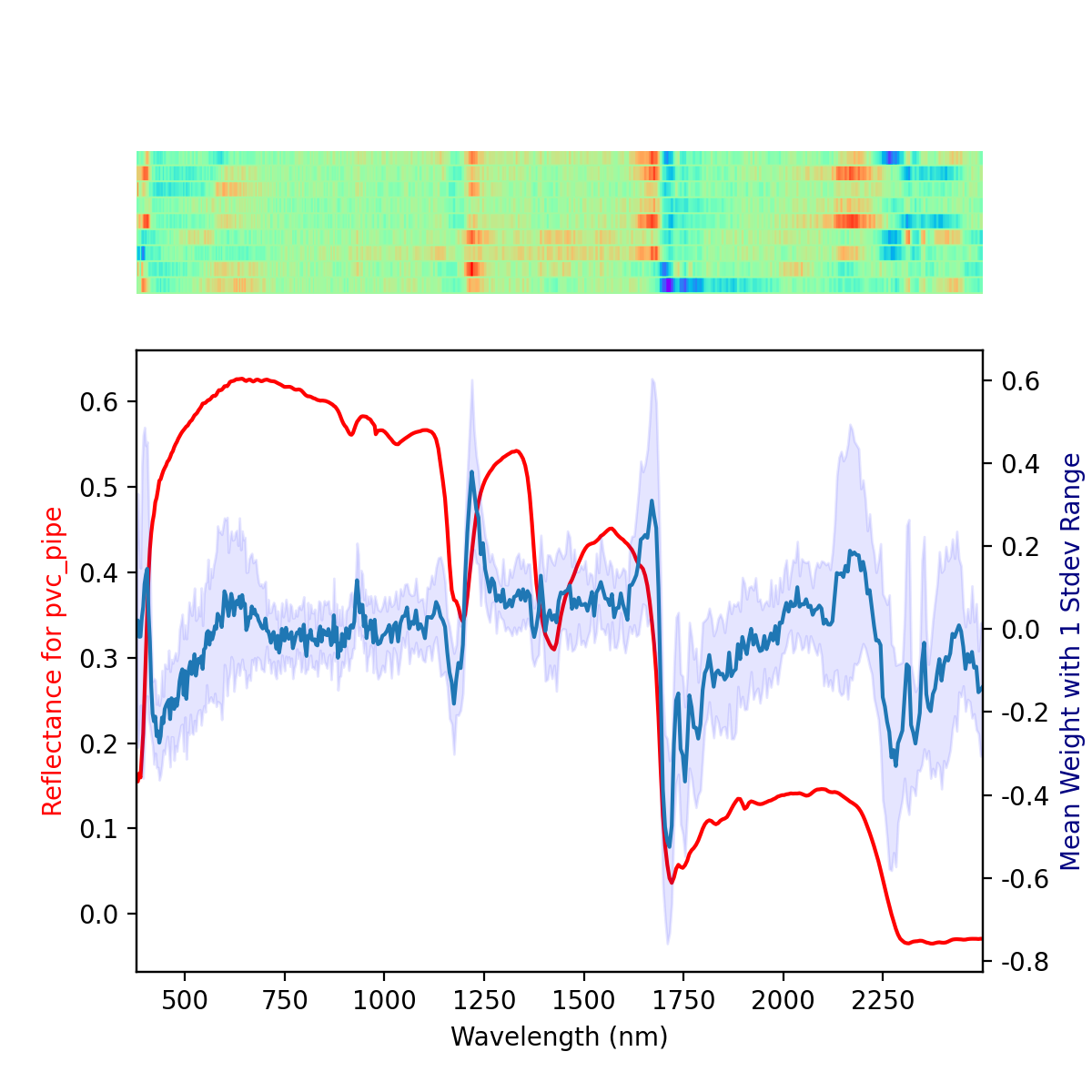}}
\vskip -0.2in
\caption{The mean reflectance for the pvc\_pipe class (red) along with the associated weights.}
\label{POLYspec5}
\end{center}
\vskip -0.3in
\end{figure}

\begin{figure}[ht]
\begin{center}
\centerline{\includegraphics[width=0.85\columnwidth]{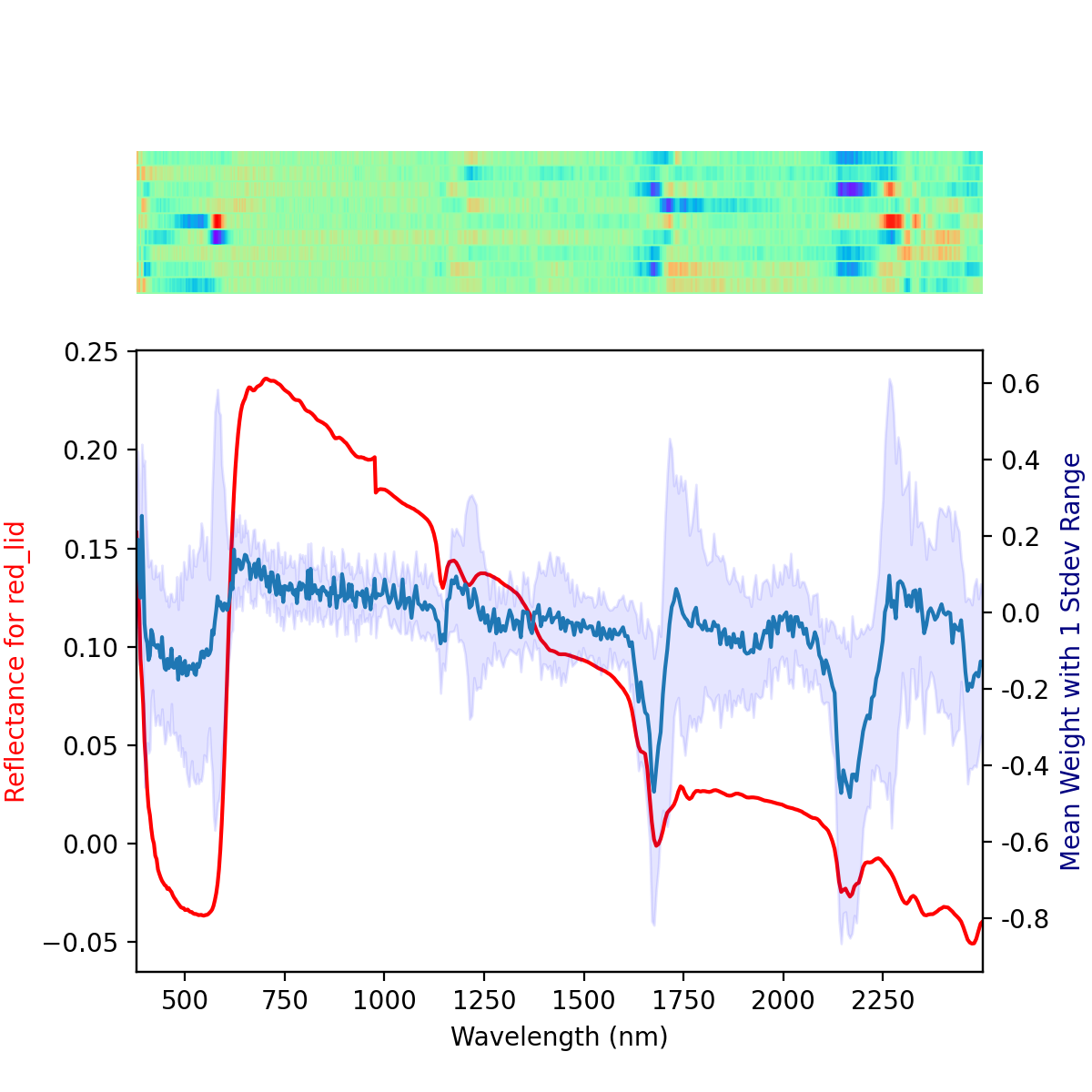}}
\vskip -0.2in
\caption{The mean reflectance for the red\_lid class (red) along with the associated weights.}
\label{POLYspec6}
\end{center}
\vskip -0.3in
\end{figure}

\begin{figure}[ht]
\begin{center}
\centerline{\includegraphics[width=0.85\columnwidth]{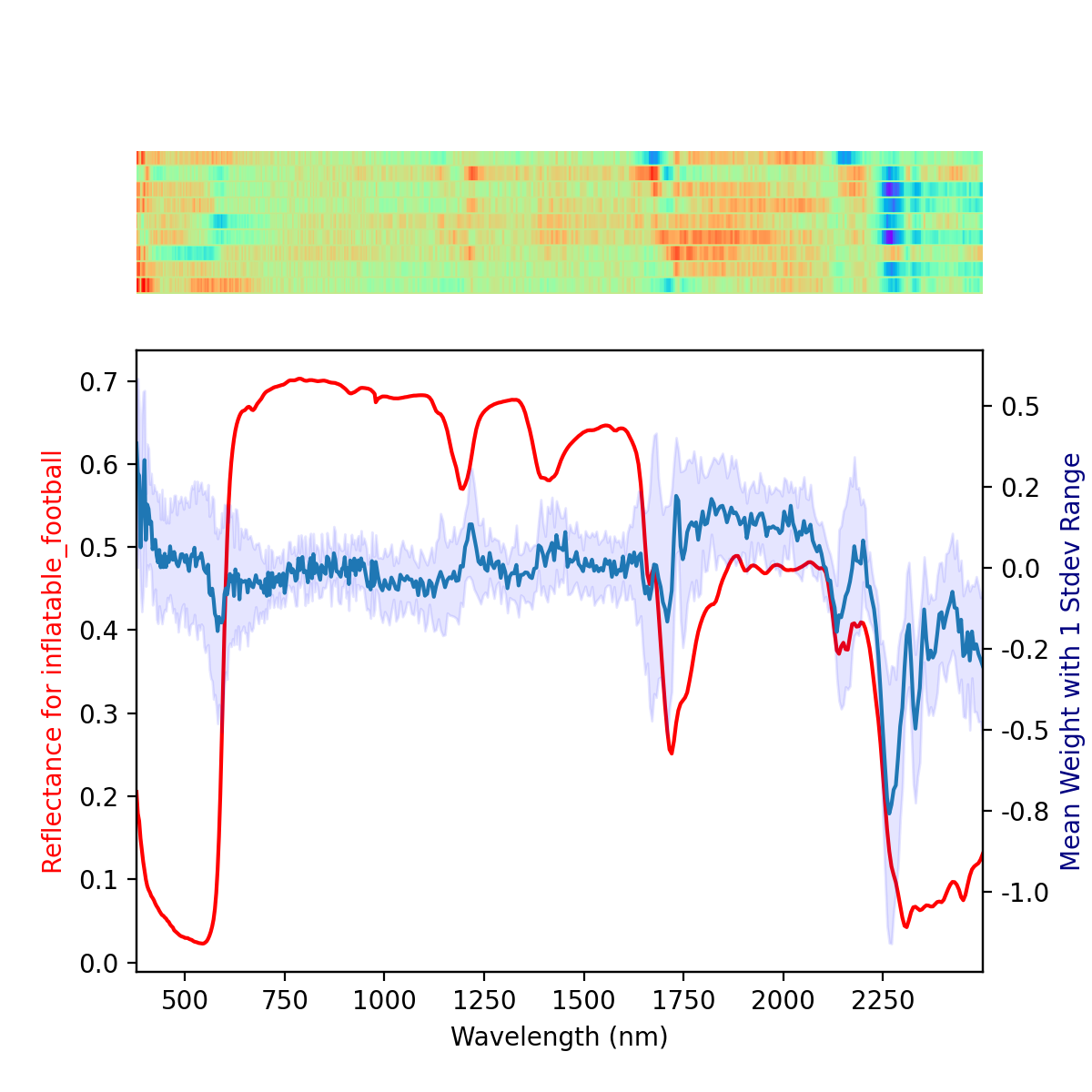}}
\vskip -0.2in
\caption{The mean reflectance for the inflatable\_football class (red) along with the associated weights.}
\label{POLYspec7}
\end{center}
\vskip -0.3in
\end{figure}

\begin{figure}[ht]
\begin{center}
\centerline{\includegraphics[width=0.85\columnwidth]{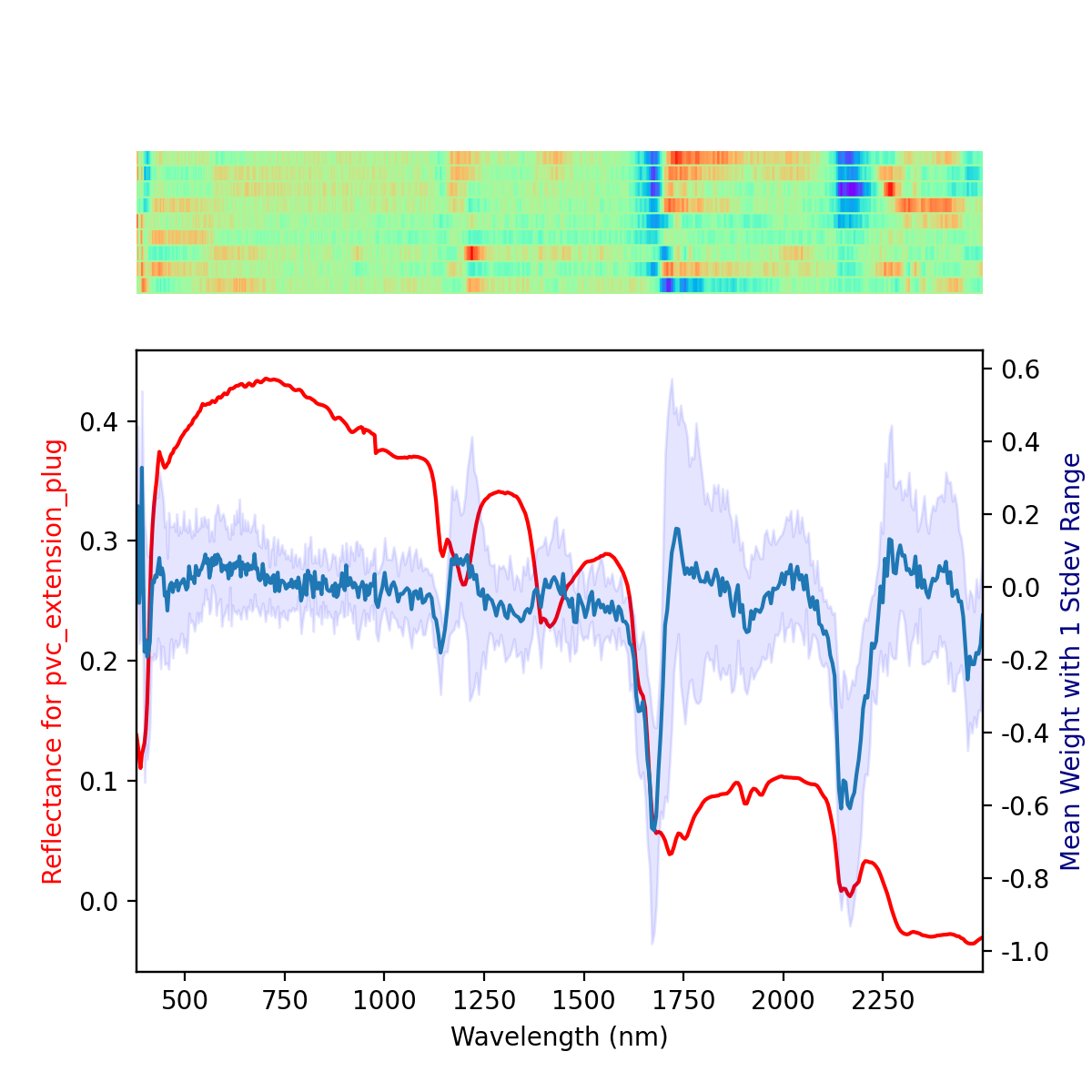}}
\vskip -0.2in
\caption{The mean reflectance for the pvc\_extension\_plug class (red) along with the associated weights.}
\label{POLYspec8}
\end{center}
\vskip -0.3in
\end{figure}

\begin{figure}[ht]
\begin{center}
\centerline{\includegraphics[width=0.85\columnwidth]{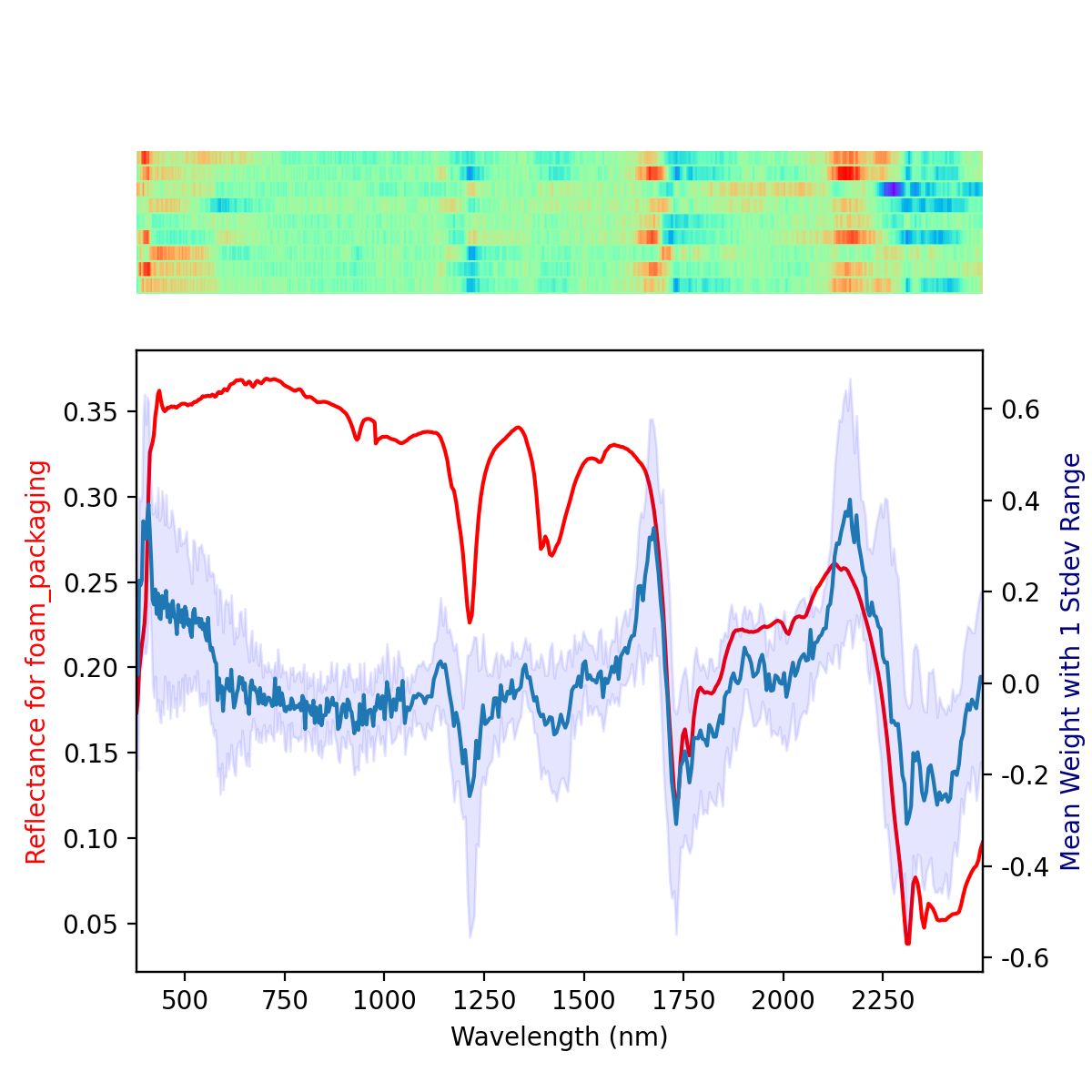}}
\vskip -0.2in
\caption{The mean reflectance for the pvc\_extension\_plug class (red) along with the associated weights.}
\label{POLYspec9}
\end{center}
\vskip -0.3in
\end{figure}

Plots of the mean reflectance spectrum for each polymer class is shown in Figures~\ref{POLYspec0} through~\ref{POLYspec9} along with the mean and standard deviation for the ten neurons most strongly associated with this class (as measured by weights from the hidden layer to the classification layer).  Above each plot is an image of the matrix of these weights aligned so the weight in the image is associated with the band in the plots directly below it.

\section{Conclusions}
We showed that in these shallow neural networks with a single hidden layer, the weights learned by the network as strongly related to features in the spectra.  These shallow networks had near 100\% accuracy on test data from a test-train split.

In the simple case of three vegetation classes with differing chlorophyll levels, the network learned a measurement similar to NDVI.  Because the weights per band seem more impactful closer to the red edge, this seems to perhaps be an interesting band-weighted form of NDVI.

For the far more complex spectra with multiple sharp features located across the spectra, the network trained on the polymer classes clearly learned which specific features were important for distinguishing between these spectra, and at the same time learned how to measure these features.  The relationship between the weights and class men spectra shown in Figure~\ref{POLYspec0} through~\ref{POLYspec9} are extraordinary.

Hopefully, this paper helps make neural networks for hyperspectral imagery more interpretable, and less of a black box method.  The accuracy even with these shallow networks is impressive.  It seems that there is great future potential for neural networks in hyperspectral imaging, but the challenge lies in inventing ways to use and train them.  Simply inserting a NN in place of a already-good target detection or linear unmixing framework is probably under-utilizing them, but for example nonlinear unmixing and intimate mixtures are a promising new avenue~\cite{Resmini2019}.  Neural networks, and deep learning in particular, is extraordinary powerful at learning to solve complex problems.  Given the complexity and variation of spectra in the world and physical interactions that can alter the measurement of spectra, framing problems that can be solved by deep learning is a major part of the frontier of future use in the field.  These new problems may be simple (for example training a NN to distinguish between multiple confuser materials as a second stage of target-identification) or may be complex with large varieties of materials and require new ways of preparing training data and displaying results.
\bibliography{my_refs2}
\bibliographystyle{IEEEbib}

\end{document}